\documentclass[letterpaper, 10 pt, conference]{ieeeconf}

\IEEEoverridecommandlockouts

\usepackage{bbm}
\usepackage{etoolbox}
\usepackage{multicol}
\usepackage{cite}
\usepackage[bookmarks=true]{hyperref}
\usepackage{graphics}
\usepackage{times}
\usepackage{amsmath}
\usepackage{amssymb}
\usepackage{xspace}
\usepackage{mathrsfs}
\usepackage{amsfonts}
\usepackage{wrapfig}

\usepackage{kantlipsum}
\usepackage{subcaption}
\usepackage{marvosym}

\usepackage{graphicx}
\usepackage{float}
\usepackage{ctable}
\usepackage{cuted}
\usepackage{colortbl}
\usepackage{multirow}
\usepackage[misc,geometry]{ifsym}
\usepackage{algorithm}
\usepackage{algpseudocode}
\usepackage[table]{xcolor}
\usepackage{pifont}
\usepackage{array}
\usepackage{tabularx}
\usepackage{adjustbox}
\let\labelindent\relax
\usepackage{enumitem}
\usepackage{makecell}

\makeatletter
\let\NAT@parse\undefined
\makeatother

\definecolor{ADHeader}{RGB}{236,239,243}
\definecolor{ADDefault}{RGB}{228,239,251}

\newcommand{\method}{\textsc{AD-WM}\xspace}
\newcommand{\sg}{\operatorname{sg}}

\title{\LARGE \bf
\method: \underline{A}ction-\underline{D}iscriminative \underline{W}orld \underline{M}odels
for Counterfactual Model Predictive Control
}

\author{%
\authorblockN{Jiabin Qiu$^{*}$, Zixuan Chen$^{*,\dagger}$, Hongye Cao,\\
Jieqi Shi$^{\dagger}$, Jing Huo$^{\dagger}$, Yang Gao}%
\authorblockA{Nanjing University\\
{\small $^{*}$Equal contribution. $^{\dagger}$Corresponding authors.}}%
}

\begin{document}
\maketitle

\thispagestyle{empty}
\pagestyle{empty}

\begin{abstract}
Latent world models are typically trained to predict factual transitions, whereas model predictive control (MPC) must compare alternative actions from the same state. A model can therefore achieve low factual prediction error yet poorly distinguish candidate actions. We introduce \method, an action-discriminative joint-embedding world model for counterfactual MPC. \method combines residual latent dynamics with predictor-level action-recovery regularization, using inverse dynamics and a normalized recovery objective motivated by conditional mutual information. Both objectives encourage planning transitions to preserve action information; their auxiliary heads are discarded at test time, leaving MPC unchanged. On OGBench-Cube, \method improves hard-start success from $3.7\%$ to $52.0\%$ over a matched LeWM baseline and improves mean success over the reproduced baseline in four of five simulation environments. Planning diagnostics show that factual prediction error and whole-bank action ranking do not follow the closed-loop success ordering, whereas CEM-aligned elite regret tracks success more closely. With a frozen V-JEPA 2 encoder and matched DROID post-training, \method also improves zero-shot transfer to our Franka setup, increasing basic pick-and-place success from $42.2\%$ to $71.1\%$ without lab-specific adaptation. These results suggest that world models for planning should preserve action-dependent differences needed for counterfactual selection, rather than optimize factual prediction accuracy alone. More videos and code are available at \href{https://ad-wm.github.io/}{https://ad-wm.github.io/}.
\end{abstract}

\section{Introduction}

World models enable agents to predict the consequences of candidate actions before executing them. For vision-based control, methods such as PlaNet, Dreamer, and TD-MPC learn latent dynamics from visual observations and use predicted trajectories to guide action selection~\cite{planet,dreamer,dreamerv2,dreamerv3,tdmpc,tdmpc2}. More recently, joint-embedding predictive architectures (JEPAs) have enabled future representation prediction without pixel reconstruction~\cite{ijepa,vjepa}. Building on this approach, DINO-WM, PLDM, LeWorldModel (LeWM), and V-JEPA~2-AC use predicted features to plan toward visual goals~\cite{dinowm,pldm,leworldmodel,vjepa2}. These methods evaluate candidate action sequences through latent rollouts and select actions whose predicted outcomes approach the goal.

\begin{figure}
    \centering
    \includegraphics[width=\linewidth]{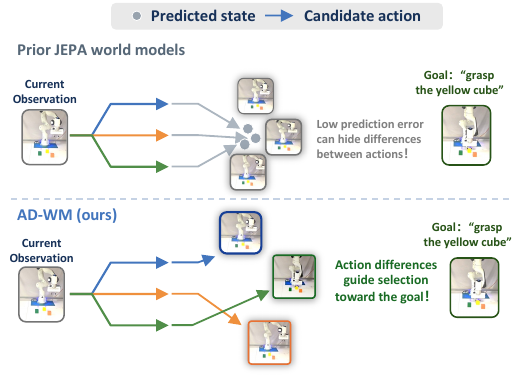}
    \caption{\textbf{Action discrimination for planning.}
    Low prediction error can mask differences between candidate actions.
    \method preserves action-dependent changes to guide goal-directed selection.}
    \vspace{-5mm}
    \label{fig:teaser}
\end{figure}

Accurate prediction of recorded transitions, however, does not guarantee useful comparisons between alternative actions. Training supervises the outcome of the action actually taken, whereas planning compares what would happen under different actions from the same state. Fig.~\ref{fig:teaser} illustrates this distinction through a cube-grasping example. When visual representations are dominated by persistent scene content, a predictor can achieve low error by largely preserving the current representation. Yet its predictions may obscure the smaller action-dependent changes that distinguish approaching the cube from moving away. Such predictions fit observed transitions but provide limited guidance for goal-directed action selection.

This motivates a key requirement for planning: \textit{latent dynamics should preserve the action-dependent differences needed for counterfactual comparison}. Factual supervision anchors predictions to observed outcomes but does not explicitly enforce action recoverability. Recovering the action from the current and predicted next representations provides a complementary training signal. Applied to model-generated transitions, this constraint directly shapes the dynamics that MPC uses to compare candidate actions. We therefore focus on \emph{predictor-level} action discrimination.

We propose \textbf{\method}, which combines residual latent prediction with predictor-level action recovery for counterfactual MPC. The residual predictor estimates latent increments relative to the current state. Action recovery uses two related objectives: inverse dynamics and normalized recovery motivated by conditional mutual information. By default, both operate on the current and predicted next representations. In simulation, the encoder and predictor are trained jointly. At deployment, the auxiliary heads are discarded, leaving MPC unchanged.

We also examine whether the learned dynamics support useful candidate selection. The cross-entropy method (CEM) retains a small set of high-scoring action sequences, called the elite set, to guide subsequent search~\cite{cem}. A model can rank most candidates correctly while misjudging those retained for further search. We therefore introduce planning-aligned diagnostics that assess selected elites using environment-realized outcomes on shared candidate sequences. In controlled Cube experiments, elite regret is more closely associated with closed-loop success than factual prediction error or whole-bank ranking. Across five simulation environments, \method improves mean success over matched LeWM in four environments. It also improves zero-shot transfer to a real robot without lab-specific adaptation.

Our contributions are threefold:
\begin{itemize}
    \item We identify a mismatch between factual prediction accuracy and counterfactual action comparison in JEPA-based MPC. Our elite-selection diagnostics better reflect closed-loop success in controlled Cube experiments.

    \item We introduce \method, which combines residual latent prediction with predictor-level action recovery to preserve action information without modifying the MPC planner.

    \item We evaluate \method across five simulation environments and a real Franka robot. Baseline comparisons, controlled ablations, and planning diagnostics establish its control benefits and examine the factors supporting transfer.
\end{itemize}

\begin{figure*}[t]
  \centering
  \includegraphics[width=\textwidth]{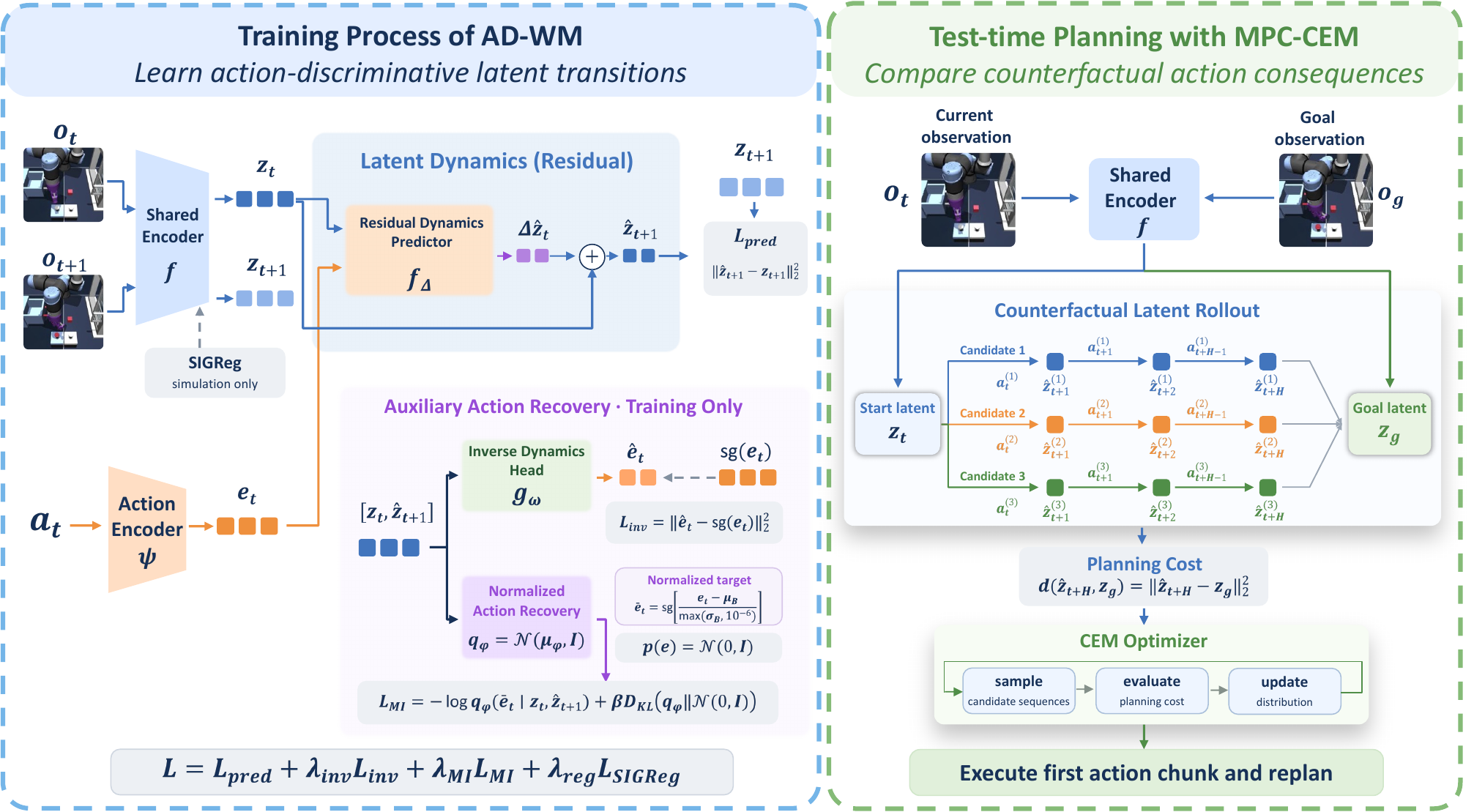}
  \caption{\textbf{Training and planning with \method.} Left: residual dynamics predicts increments supervised by encoded successors. Inv and MI encourage action recovery from predicted transitions. Simulation jointly trains the encoder and predictor with SIGReg. Robot post-training freezes the encoder and omits SIGReg. Right: CEM scores rollouts by terminal cost $\|\hat z_{t+H}-z_g\|_2^2$, executes the first action block, and replans. Auxiliary heads are discarded at deployment.}
  \label{fig:method}
  \vspace{-5mm}
\end{figure*}

\section{Related Work}

\textbf{Latent World Models for Control.}
Learned dynamics support model-based control by predicting the consequences of actions before execution~\cite{worldmodels,sutton1991dyna}. Existing approaches use these predictions for latent-space planning, behavior learning through imagined rollouts, or visual foresight~\cite{planet,dreamer,tdmpc,e2c,visual_foresight}. More recent work explores Transformer and diffusion architectures for world modeling~\cite{transformer_world_models,diffusion_world_model}. Robot world models also investigate action-conditioned prediction and counterfactual behavior~\cite{dreamdojo}. These approaches differ in how their predictions guide control. Our work focuses on reward-free, image-goal MPC learned from offline data. In this setting, predicted latent transitions directly determine how the planner compares candidate action sequences. We therefore study whether these transitions preserve the action-dependent differences needed for selection.

\textbf{Feature-predictive world models.}
Joint-embedding predictive architectures (JEPAs) predict future representations without explicit observation reconstruction~\cite{ijepa,vjepa}. DINO-WM, PLDM, LeWM, and V-JEPA~2 use learned or pretrained visual features for goal-directed planning~\cite{dinowm,pldm,leworldmodel,vjepa2}. LeWM is our closest end-to-end simulation baseline, while V-JEPA~2-AC provides pretrained dynamics for our real-robot study. JEPA's emphasis on slowly varying features~\cite{slow_features} raises the question of whether predictions adequately capture smaller action-dependent changes needed for control.
Recent extensions address complementary limitations of JEPA-based control. Fast-LeWM improves rollout efficiency, while Sub-JEPA regularizes representations for stable end-to-end learning~\cite{gao2026fast,zhao2026subjepa}. INTACT and Qantara introduce action-generation and control interfaces beyond standard CEM planning~\cite{sun2026intact,qantara2026}. Delta-JEPA promotes action sensitivity by recovering actions from observed latent displacements~\cite{zhang2026deltajepa}. Our normalized recovery regularizes predicted transitions, as does the default inverse objective. Combined with residual prediction, these objectives shape latent dynamics for counterfactual action comparison without modifying the MPC planner.

\textbf{Action-aware representation learning.}
Inverse dynamics provides a training signal for learning action-relevant representations from observed state transitions~\cite{learning_to_poke,pathak_inverse_features}. Related work studies Markov state abstractions and action-sufficient representations for control~\cite{markov_abstraction,action_sufficient}. These approaches motivate retaining information that is useful for predicting and influencing environment dynamics. We apply this principle directly at the predictor level. Our inverse dynamics and normalized action-recovery objectives operate on the current and predicted next representations. Their gradients therefore constrain the model-generated transitions that MPC rolls out and compares during planning. This complements factual prediction supervision with an explicit objective for preserving action information in predicted transitions.

\section{Method}
\label{sec:approach}

Fig.~\ref{fig:method} illustrates the training and planning pipeline of \method. During training, residual latent prediction and action-recovery objectives jointly shape action-discriminative transitions. At test time, MPC uses the learned dynamics to compare candidate action sequences, with the auxiliary heads discarded. We first formulate the task, then detail the model, training objectives, and planning procedure. Finally, we introduce diagnostics for evaluating candidate selection.

\subsection{Problem Formulation}
\label{sec:problem}

Given an offline dataset $\mathcal D=\{(o_t,a_t,o_{t+1})\}$ of images and continuous actions, we learn an encoder $z_t=e_\phi(o_t)$ and latent dynamics $\hat z_{t+1}=F_\theta(z_t,a_t)$. At test time, the agent receives a current image and a goal image $g$, encoded as $z_g=e_\phi(g)$. MPC evaluates candidate action sequences $a=a_{t:t+H-1}$ using the predicted terminal cost
$\hat c(a)=\|\hat z_{t+H}(z_t,a)-z_g\|_2^2$.
It executes the first action block and replans without parameter updates.
Low factual prediction error alone does not ensure useful action comparison. For transitions $z_{t+1}=z_t+\delta_t$ with small increments, an action-independent predictor $F_0(z,a)=z$ incurs only $\mathbb E\|\delta_t\|_2^2$ expected error. Yet it assigns all action sequences the same terminal cost. We therefore seek latent dynamics that preserve action-dependent differences needed for planning.

\subsection{Residual Latent Prediction}

Successive visual representations often share substantial state information. We let the current representation carry this shared information and train the predictor to estimate the remaining change. Given a learned action embedding $e_t=\psi_\rho(a_t)$, \method predicts:
\begin{equation}
 \Delta\hat z_t=f_\theta(z_t,e_t),\qquad
 \hat z_{t+1}=z_t+\Delta\hat z_t.
 \label{eq:residual}
\end{equation}
The prediction is supervised by the encoded successor:
\begin{equation}
 \mathcal L_{\mathrm{pred}}=\|\hat z_{t+1}-z_{t+1}\|_2^2,
 \qquad z_{t+1}=e_\phi(o_{t+1}).
 \label{eq:pred}
\end{equation}
This equivalently matches $\Delta\hat z_t$ to the encoded displacement $z_{t+1}-z_t$. Absolute and residual prediction share the same optimum in an unconstrained function class. The residual parameterization changes the learning bias by explicitly modeling local latent changes. However, it does not itself require these changes to retain action information. We address this through predictor-level action recovery.

\subsection{Predictor-Level Action Recovery}

We encourage actions to be recoverable from the current and predicted next representations. Applying recovery to model-generated transitions directly shapes the dynamics used by MPC. Prediction supervision anchors these transitions to observed outcomes, while recovery discourages the predictor from ignoring action information.

\textbf{Inverse dynamics.}
The default inverse head recovers the action embedding from predicted transition endpoints:
\begin{equation}
 \hat e_t=g_\omega(z_t,\hat z_{t+1}),\qquad
 \mathcal L_{\mathrm{inv}}=\|\hat e_t-\sg(e_t)\|_2^2.
 \label{eq:inv}
\end{equation}
The operator $\sg$ detaches the recovery target. The predicted endpoint remains differentiable, allowing auxiliary gradients to train the predictor and action encoder $\psi_\rho$ through $\hat z_{t+1}$. Thus, recovery constrains generated transitions rather than only observed state representations. Section~\ref{sec:ablation} examines alternative inverse inputs.

\textbf{Normalized action recovery.}
Inv operates on learned action embeddings at their current scale. Normalized recovery uses standardized targets and a Gaussian recovery objective motivated by conditional mutual information (MI). We normalize embeddings componentwise as
$\bar e_t=\sg[(e_t-\mu_B)/\max(\sigma_B,10^{-6})]$,
where $\mu_B$ and $\sigma_B$ are the batch mean and standard deviation. This controls target scale for likelihood training with fixed unit covariance.

A separate head receives $[z_t,\hat z_{t+1}]$ and defines
$q_\eta(\cdot\mid z_t,\hat z_{t+1})=\mathcal N(\mu_\eta,I)$.
Using a standard Gaussian as a fixed regularization reference, we minimize:
\begin{equation}
\begin{split}
 \mathcal L_{\mathrm{MI}}={}&-\log q_\eta(\bar e_t\mid z_t,\hat z_{t+1})\\
 &+\beta D_{\mathrm{KL}}\!\left(q_\eta(\cdot\mid z_t,\hat z_{t+1})\,\|\,\mathcal N(0,I)\right).
\end{split}
\label{eq:mi}
\end{equation}
With unit covariance, this reduces, up to an additive constant, to:
\begin{equation*}
 \tfrac12\|\bar e_t-\mu_\eta\|_2^2
 +\tfrac\beta2\|\mu_\eta\|_2^2.
\end{equation*}
The first term recovers standardized action embeddings. The second shrinks the predicted mean toward zero. For fixed state/action representations and target normalization, the recovery log-likelihood is the variational term in a Barber--Agakov lower bound on $I(\bar e_t;\hat z_{t+1}\mid z_t)$~\cite{barber2003information}. This motivates the recovery term, with additional KL regularization during joint training.

\subsection{Training and Planning}

In simulation, the encoder and predictor are trained jointly with:
\begin{equation}
 \mathcal L=\mathcal L_{\mathrm{pred}}+\lambda_{\mathrm{sig}}\mathcal L_{\mathrm{sig}}
 +\lambda_{\mathrm{inv}}\mathcal L_{\mathrm{inv}}
 +\lambda_{\mathrm{MI}}\mathcal L_{\mathrm{MI}}.
 \label{eq:full}
\end{equation}
Prediction supervision matches observed outcomes, while the recovery objectives encourage predicted transitions to retain action information. We use the same SIGReg representation regularizer as LeWM~\cite{leworldmodel}. Robot post-training freezes the encoder and omits SIGReg (Section~\ref{sec:robot}).

At deployment, both auxiliary heads are discarded. MPC embeds candidate actions with $\psi_\rho$ and recursively applies Eq.~\eqref{eq:residual}. CEM retains the sequences with lowest predicted terminal cost and refits its sampling distribution to this elite set. After search, the agent executes the first action block and replans from the new observation. The encoder architecture and CEM procedure remain unchanged.

\begin{table*}[t!]
\centering
\caption{\textbf{Cube success (\%) on matched starts.} Fast-LeWM and Sub-JEPA use one checkpoint. Others report mean$\pm$population SD over three. External methods retain native inference. Bold marks the highest mean.}
\label{tab:external-hardstart}
\footnotesize
\setlength{\tabcolsep}{3pt}
\renewcommand{\arraystretch}{1.02}
\begin{tabular*}{\textwidth}{@{\extracolsep{\fill}}llrrrrrr@{}}
\toprule
\rowcolor{ADHeader}
Method & Inference & Original & P00 & P01 & P02 & P03 & P04 \\
\midrule
LeWM & CEM & $73.3{\pm}2.5$ & $8.0{\pm}2.8$ & $4.7{\pm}0.9$ & $4.7{\pm}1.9$ & $1.3{\pm}1.9$ & $0.0{\pm}0.0$ \\
Fast-LeWM~\cite{gao2026fast} & CEM & 80.0 & 24.0 & 20.0 & 10.0 & 2.0 & 0.0 \\
Sub-JEPA~\cite{zhao2026subjepa} & CEM & 78.0 & 34.0 & 20.0 & 10.0 & 10.0 & 10.0 \\
INTACT~\cite{sun2026intact} & Pure CEM & $74.7{\pm}0.9$ & $20.0{\pm}3.3$ & $21.3{\pm}9.3$ & $13.3{\pm}2.5$ & $6.7{\pm}1.9$ & $3.3{\pm}2.5$ \\
INTACT & Actor-CEM & $94.7{\pm}2.5$ & $78.7{\pm}3.4$ & $72.0{\pm}5.7$ & $56.0{\pm}4.3$ & $33.3{\pm}12.4$ & $10.7{\pm}2.5$ \\
INTACT & Direct & $\mathbf{98.7{\pm}1.9}$ & $\mathbf{99.3{\pm}0.9}$ & $\mathbf{99.3{\pm}0.9}$ & $\mathbf{88.7{\pm}1.9}$ & $35.3{\pm}3.4$ & $10.0{\pm}0.0$ \\
\midrule
\rowcolor{ADDefault}
\textbf{\method} & CEM & $90.7{\pm}3.4$ & $74.0{\pm}2.8$ & $68.0{\pm}4.3$ & $56.0{\pm}1.6$ & $\mathbf{36.7{\pm}5.2}$ & $\mathbf{25.3{\pm}3.4}$ \\
\bottomrule
\end{tabular*}
\vspace{-5mm}
\end{table*}

\subsection{Planning-Facing Diagnostics}
\label{sec:planning_diagnostics}

To assess whether predicted transitions support useful action selection, we evaluate global ranking and elite quality on a shared candidate bank $\mathcal A=\{a^{(i)}\}_{i=1}^N$. For each sequence, $\hat c(a)$ is its predicted terminal cost. Its realized cost $c^\star(a)$ is obtained by executing the sequence from the same initial environment state and encoding the final image. These realized costs are used only for simulation diagnostics, outside training and MPC. Each model uses its own encoder for both costs, so the metrics assess alignment within its planning space.

\textbf{Global ranking.}
Counterfactual action discriminability (CAD) measures agreement across the full candidate bank:
\begin{equation}
 \mathrm{CAD}(z_t)=\rho_S\!\left(
 \{\hat c(a^{(i)})\}_{i=1}^N,
 \{c^\star(a^{(i)})\}_{i=1}^N\right),
 \label{eq:cad}
\end{equation}
where $\rho_S$ is Spearman rank correlation. This includes candidates outside the elite set.

\textbf{Elite quality.}
Let $E_k(c)$ denote the $k$ lowest-cost candidates under $c$. Define the realized cost range
$D_{\mathcal A}=\max_{a\in\mathcal A}c^\star(a)-\min_{a\in\mathcal A}c^\star(a)$,
floored at $10^{-8}$. Normalized \emph{best-in-elite regret} is:
\begin{equation}
 R_k=\frac{\min_{a\in E_k(\hat c)}c^\star(a)-\min_{a\in\mathcal A}c^\star(a)}
 {D_{\mathcal A}}.
 \label{eq:regret}
\end{equation}
Low $R_k$ indicates that the predicted elite set retains at least one near-best candidate. Since CEM refits its proposal using all elites, we also define \emph{elite-mean regret}:
\begin{equation}
 \bar R_k=\frac{\sum_{a\in E_k(\hat c)}c^\star(a)-\sum_{a\in E_k(c^\star)}c^\star(a)}
 {kD_{\mathcal A}}.
 \label{eq:mean_regret}
\end{equation}
This compares the mean realized cost of predicted elites with that of the realized top-$k$ set. Both regrets normalize selection penalties by the bank's realized cost range. They characterize candidate selection, while closed-loop success remains the final control outcome.

\section{Experiments}
\label{sec:experiments}

We organize our experiments around four questions: \textbf{Q1:} How well does \method perform across simulation tasks and configuration shifts? \textbf{Q2:} How do residual prediction and action recovery contribute to control performance?
\textbf{Q3:} Which prediction and selection metrics best reflect closed-loop success?
\textbf{Q4:} Can \method transfer to a real robot without lab-specific adaptation?

\subsection{Experimental Setup}

\begin{figure}
    \centering
    \includegraphics[width=\linewidth]{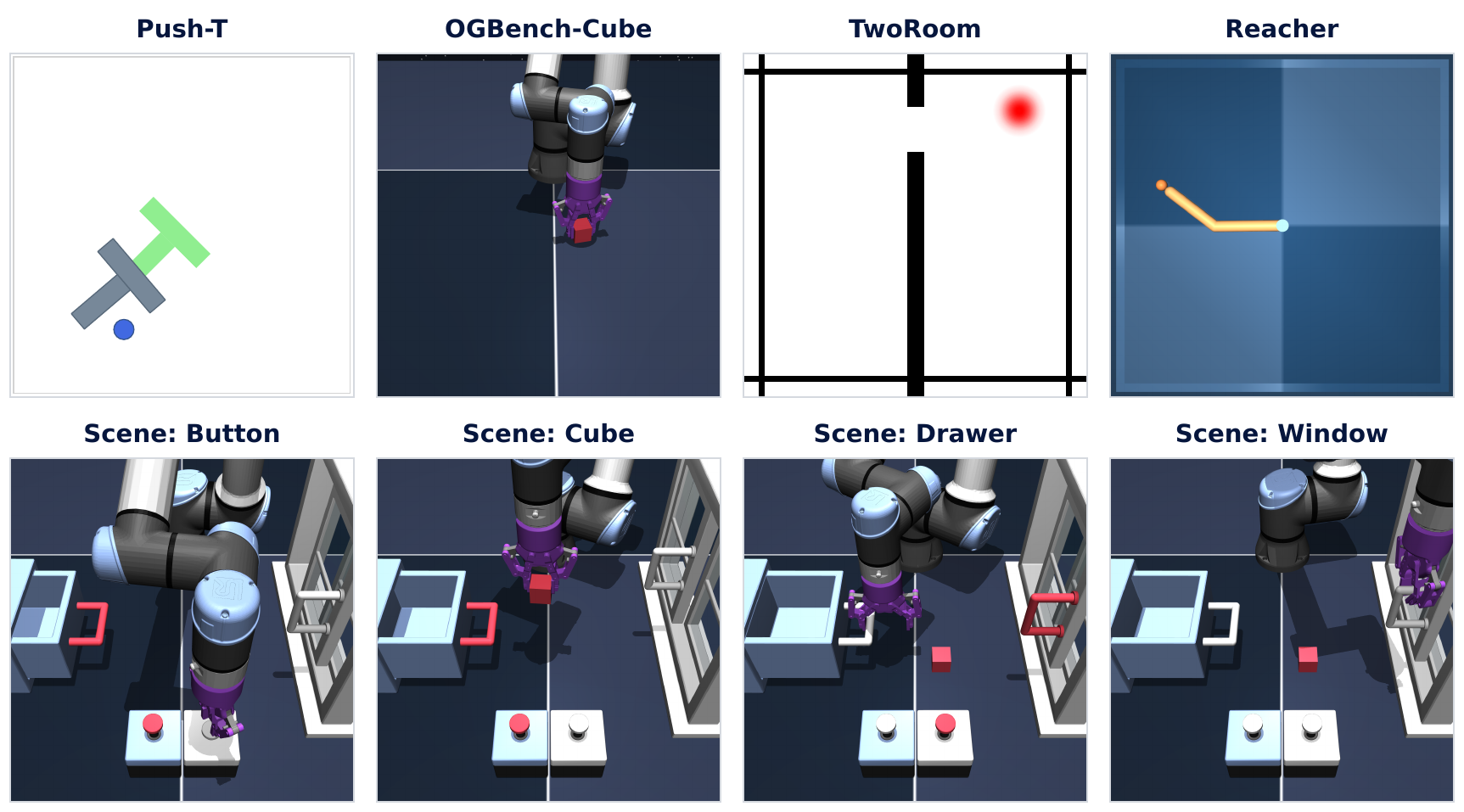}
    \caption{The simulation environments.}
    \label{fig:sim_env}
    \vspace{-5mm}
\end{figure}

\textbf{Benchmarks and Baselines.}
We follow LeWM's simulation environment selection~\cite{leworldmodel}, evaluating on Cube, Reacher, TwoRoom, and PushT, and additionally include Scene from OGBench~\cite{ogbench}. As shown in Fig.~\ref{fig:sim_env}, these environments cover robotic manipulation, reaching, navigation, and planar pushing. Cube serves as the primary benchmark for controlled ablations and planning diagnostics. We retain the original evaluation protocols for cross-environment comparisons, except for Scene, and introduce additional hard-start protocols on Cube.
Our main baseline is a matched LeWM reproduction~\cite{leworldmodel}. We also evaluate released Cube checkpoints from Fast-LeWM~\cite{gao2026fast}, Sub-JEPA~\cite{zhao2026subjepa}, and INTACT~\cite{sun2026intact}, retaining their native inference interfaces. Controlled ablations share images, encoder size, training budget, and MPC settings, varying only transition parameterization and auxiliary training choices.

\textbf{Evaluation protocols.}
Cube includes Original, which samples dataset states, and hard-start protocols P00--P04. Hard starts select tabletop states without gripper contact, with minimum goal distance $0.05$ and end-effector--cube distance $0.02$. P00--P04 apply cube $xy$ perturbations of radii $0.00/0.01/0.02/0.03/0.04$, clipped to $x\in[0.30,0.55]$ and $y\in[-0.30,0.30]$. P00 therefore differs from Original even without perturbation. Each protocol uses 50 episodes per checkpoint with evaluation seed 42. Hard-start success (HS) averages P00--P04. Unless stated otherwise, Cube results are means and population standard deviations (SDs) over seeds 3072, 4096, and 6144.
Reacher, TwoRoom, and PushT use original protocols. Scene uses 200 hard-start episodes per checkpoint, balanced across Button, Cube, Drawer, and Window. Starts differ from goals in one component, with change-detection tolerance $0.04$ and minimum target displacement and end-effector--target distance $0.08$ for geometric filters.

We report success as percentages and factual latent mean squared error (MSE) at one step and averaged over the first five rollout steps (local MSE). Counterfactual action discriminability (CAD) measures whole-bank ranking agreement; $R_{30}$ and $\bar R_{30}$ denote normalized best-in-elite and elite-mean regret for 30 selected candidates (Section~\ref{sec:planning_diagnostics}). We use $\rho$ for Spearman rank correlation.

\textbf{Training details.}
Our simulation setup follows LeWM~\cite{leworldmodel}, using $224\times224$ images, training/evaluation histories of 3/1, and frame skip 5. The trainable ViT-tiny encoder~\cite{vit} has latent dimension 192. The predictor has depth 6, 16 heads, and MLP dimension 2048. Inv and MI heads use 1024-hidden-unit MLPs with BatchNorm. Training uses bf16, 10 epochs, AdamW, batch size 128, learning rate $5\times10^{-5}$, weight decay $10^{-3}$, and $\lambda_{\mathrm{sig}}=0.09$. All simulation environments use $\lambda_{\mathrm{inv}}=0.1$, $\lambda_{\mathrm{MI}}=0.01$, and $\beta=0.01$, except Scene, where $\lambda_{\mathrm{MI}}=10^{-4}$. Primary weights were specified before sensitivity analysis.

\textbf{Planning details.}
Matched comparisons use CEM~\cite{cem} with 300 candidates, 30 elites, 30 iterations, horizon 5, receding horizon 5, and action block 5. Actions are clipped to environment bounds and scored by terminal latent distance. Cube and Scene use goal offset 25 and a 50-step budget. Sub-JEPA and INTACT CEM modes use the same candidate count, elite count, iterations, horizon, and action block. Fast-LeWM uses horizon 1 and action block 25. INTACT Direct performs no search. External Cube comparisons share starts, episode counts, goal offset, and budget, but do not control training and inference as the LeWM-\method pair does.

\begin{table}[t]
\centering
\caption{\textbf{Cube ablations (\%, three seeds)}. A/B show mean$\pm$population SD, C/D show means. HS averages P00--P04. Blue marks proposed/default settings in A/C/D. Bold indicates column-best means in A and row-best means in C/D.}
\label{tab:ablations}
\label{tab:cube}
\label{tab:controls}
\footnotesize
\setlength{\tabcolsep}{2.5pt}
\renewcommand{\arraystretch}{1.02}

\begin{tabular*}{\columnwidth}{@{\extracolsep{\fill}}lrr@{}}
\toprule
\rowcolor{ADHeader}
\multicolumn{3}{l}{\textbf{A. Component contributions}} \\
Variant & Original $\uparrow$ & HS $\uparrow$ \\
\midrule
LeWM & $73.3\pm2.5$ & $3.7\pm1.4$ \\
Abs. + Inv + MI & $83.3\pm0.9$ & $14.4\pm2.9$ \\
Res & $82.7\pm2.5$ & $34.7\pm1.6$ \\
Res + Inv & $83.3\pm1.9$ & $37.1\pm4.7$ \\
Res + MI & $89.3\pm3.8$ & $\mathbf{54.7\pm3.0}$ \\
\rowcolor{ADDefault}
\textbf{\method} & $\mathbf{90.7\pm3.4}$ & $52.0\pm3.1$ \\
\bottomrule
\end{tabular*}

\par\smallskip

\begin{tabular*}{\columnwidth}{@{\extracolsep{\fill}}lrr@{}}
\toprule
\rowcolor{ADHeader}
\multicolumn{3}{l}{\textbf{B. MI gains across inverse inputs (HS)}} \\
Input & No MI & MI $=0.01$ \\
\midrule
Predicted endpoints & $37.1\pm4.7$ & $52.0\pm3.1$ \\
Encoded endpoints & $45.1\pm4.4$ & $60.7\pm0.4$ \\
Predicted increment & $39.9\pm2.1$ & $60.3\pm3.7$ \\
\bottomrule
\end{tabular*}

\par\smallskip

\begin{tabular*}{\columnwidth}{@{\extracolsep{\fill}}lrrrrrr@{}}
\toprule
\rowcolor{ADHeader}
\multicolumn{7}{l}{\textbf{C. MI weight}
($\lambda_{\mathrm{inv}}=0.1$)} \\
$\lambda_{\mathrm{MI}}$
& 0 & .005 & \cellcolor{ADDefault}.01 & .015 & .03 & .05 \\
\midrule
Original
& 83.3 & 89.3 & \cellcolor{ADDefault}90.7
& 90.7 & \textbf{91.3} & 88.0 \\
HS
& 37.1 & 50.9 & \cellcolor{ADDefault}52.0
& 60.4 & \textbf{65.2} & 61.5 \\
\bottomrule
\end{tabular*}

\par\smallskip

\begin{tabular*}{\columnwidth}{@{\extracolsep{\fill}}lrrrr@{}}
\toprule
\rowcolor{ADHeader}
\multicolumn{5}{l}{\textbf{D. Inv weight}
($\lambda_{\mathrm{MI}}=0.01$)} \\
$\lambda_{\mathrm{inv}}$
& 0 & .05 & \cellcolor{ADDefault}.10 & .20 \\
\midrule
Original
& 89.3 & 90.0 & \cellcolor{ADDefault}90.7 & \textbf{92.0} \\
HS
& 54.7 & 57.2 & \cellcolor{ADDefault}52.0 & \textbf{57.5} \\
\bottomrule
\end{tabular*}
\par\smallskip
\parbox{\columnwidth}{\scriptsize
MI-enabled variants in A use $\lambda_{\mathrm{MI}}=0.01$.
In B, $\lambda_{\mathrm{inv}}=0.1$ and MI always uses predicted endpoints. Encoded endpoints
remove only the direct Inv gradient through the predicted successor.}
\vspace{-5mm}
\end{table}

\subsection{Simulation Results}

\textbf{Main results.}
To answer Q1, Table~\ref{tab:external-hardstart} compares methods on identical Cube starts. \method outperforms matched LeWM across all protocols. INTACT Direct performs best from Original through P02, while its Actor-CEM mode improves over Pure CEM through P03. On P03, INTACT Direct, Actor-CEM, and \method have similar means relative to checkpoint variation. On P04, \method reaches $25.3\%$, compared with $10.0\%$ and $10.7\%$ for Direct and Actor-CEM. These results indicate stronger performance under larger perturbations, but differences in training and inference prevent attributing the crossover to search strategy alone.

\textbf{Configuration shift.}
P00--P01 serve as near-distribution controls, while P02--P04 probe configuration tails within empirical cube $xy$ support. We quantify shift using Euclidean nearest-neighbor distances to training states from other episodes. Joint coordinates concatenate cube $xy$ and end-effector $xyz$, while relative coordinates use end-effector-minus-cube $xyz$. Reference 95th-percentile thresholds use 20,000 tabletop/non-contact training states per seed. For P02--P04, $2.0\%/38.7\%/72.0\%$ exceed the joint threshold, and $52.7\%/80.7\%/95.3\%$ exceed the relative threshold.

\begin{figure}[t]
  \centering
  \begin{subfigure}{\columnwidth}
    \centering
    \includegraphics[width=\linewidth]{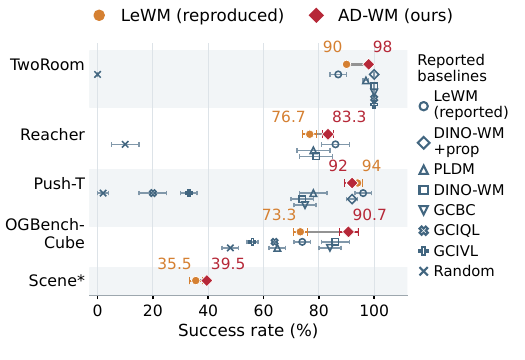}
    \caption{Cross-environment performance.}
    \label{fig:crossenv_overall}
  \end{subfigure}

  \smallskip
  \begin{subfigure}{\columnwidth}
    \centering
    \includegraphics[width=\linewidth]{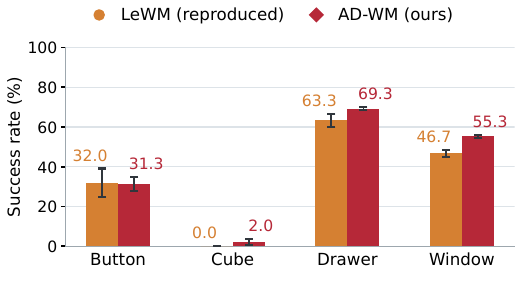}
    \caption{Scene subtask performance.}
    \label{fig:scene_subtasks}
  \end{subfigure}

  \caption{\textbf{Cross-environment and subtask performance.}
  (a) Matched LeWM and \method results, with reported
  baselines shown separately. Scene (*)
  uses balanced hard starts, while other environments use
  original protocols.
  (b) Scene results over three seeds and 50 episodes per
  subtask, with $\lambda_{\mathrm{MI}}=10^{-4}$.
  Error bars show source SDs in (a) and population SDs in (b).}
  \label{fig:crossenv}
  \vspace{-5mm}
\end{figure}

\textbf{Cross-environment performance.}
Fig.~\ref{fig:crossenv}(a) shows higher mean success than reproduced LeWM in four of five environments: Cube ($73.3\%\to90.7\%$), Reacher ($76.7\%\to83.3\%$), TwoRoom ($90\%\to98\%$), and Scene ($35.5\%\to39.5\%$). PushT decreases from $94\%$ to $92\%$. All environments except Scene use original protocols. Reported external results are shown separately, including LeWM's reported $86\%$ on Reacher. Our improvement claim concerns the matched reproduction.
Scene gains concentrate in Drawer ($63.3\%\to69.3\%$) and Window ($46.7\%\to55.3\%$), both improving in all three matched seed comparisons. Button is similar and Cube remains difficult. The overall gain remains unresolved by the paired seed test ($p=0.13$).

\subsection{Ablation Studies}
\label{sec:ablation}

Table~\ref{tab:ablations} addresses Q2 through component ablations (A), MI gains across inverse inputs (B), and post-hoc weight sensitivity (C/D).

\textbf{Component contributions.}
Panel A identifies residual prediction and MI as the largest contributors. LeWM achieves $3.7\%$ HS. Adding Inv+MI to absolute prediction raises this to $14.4\%$, while residual prediction alone reaches $34.7\%$. Res+Inv achieves $37.1\%$, Res+MI $54.7\%$, and \method $52.0\%$. Inv gives a smaller mean gain over Res and does not improve upon Res+MI at its default weight.

\textbf{MI gains across inverse inputs.}
Panel B varies Inv inputs while MI always uses predicted endpoints. MI adds $14.9/15.6/20.4$ mean HS points across predicted endpoints, encoded endpoints, and predicted increments. The latter two reach $60.7/60.3\%$, versus $52.0\%$ for the default; predicted increments still backpropagate Inv through the predictor. Thus, B tests Inv-input robustness, not MI-input robustness.

\textbf{Weight sensitivity.}
With Inv fixed at $0.1$, all tested nonzero MI weights ($0.005$--$0.05$) yield higher mean HS: $50.9$--$65.2\%$ versus $37.1\%$ (Panel C). MI peaks at $0.03$; the pre-specified $0.01$ remains primary. With MI fixed at $0.01$, Inv weights $0.05/0.20$ exceed the no-Inv mean, but the default $0.10$ does not (Panel D). Inv's effect is smaller and weight-dependent.

\begin{table*}[t]
\centering
\caption{\textbf{Prediction, selection, and control metrics on Cube (mean$\pm$population SD, three seeds)}. Local MSE averages five rollout steps. MSE values are multiplied by $10^3$. Shared banks use $N=300$ and $k=30$. Blue marks \method. Bold marks the best displayed mean per metric, including ties.}
\label{tab:diagnostic-summary}
\label{tab:factual}
\label{tab:commonbank}
\footnotesize
\setlength{\tabcolsep}{3pt}
\renewcommand{\arraystretch}{1.05}
\begin{tabular*}{\textwidth}{@{\extracolsep{\fill}}lrrrrrr@{}}
\toprule
\rowcolor{ADHeader}
Variant & \multicolumn{2}{c}{Factual prediction} & \multicolumn{3}{c}{Candidate selection} & Control \\
\cmidrule(lr){2-3}\cmidrule(lr){4-6}\cmidrule(lr){7-7}
 & One-step MSE $\downarrow$ & Local MSE $\downarrow$ & CAD $\uparrow$ & $R_{30}\downarrow$ & $\bar R_{30}\downarrow$ & HS (\%) $\uparrow$ \\
\midrule
LeWM & $\mathbf{2.72\pm0.06}$ & $\mathbf{10.36\pm0.48}$ & $0.460\pm0.010$ & $0.074\pm0.006$ & $0.298\pm0.010$ & $3.7\pm1.4$ \\
Res & $3.61\pm0.20$ & $11.17\pm0.61$ & $0.469\pm0.005$ & $0.046\pm0.003$ & $0.250\pm0.009$ & $34.7\pm1.6$ \\
Res + Inv & $3.43\pm0.11$ & $11.00\pm0.28$ & $\mathbf{0.470\pm0.005}$ & $0.043\pm0.002$ & $0.245\pm0.005$ & $37.1\pm4.7$ \\
Res + MI & $4.20\pm0.06$ & $15.17\pm0.04$ & $0.454\pm0.002$ & $\mathbf{0.029\pm0.002}$ & $\mathbf{0.228\pm0.002}$ & $\mathbf{54.7\pm3.0}$ \\
\rowcolor{ADDefault}
\textbf{\method} & $4.27\pm0.10$ & $15.77\pm0.20$ & $0.448\pm0.011$ & $\mathbf{0.029\pm0.001}$ & $0.229\pm0.005$ & $52.0\pm3.1$ \\
\bottomrule
\end{tabular*}
\end{table*}

\begin{figure*}[t]
  \centering
  \includegraphics[width=0.95\textwidth]{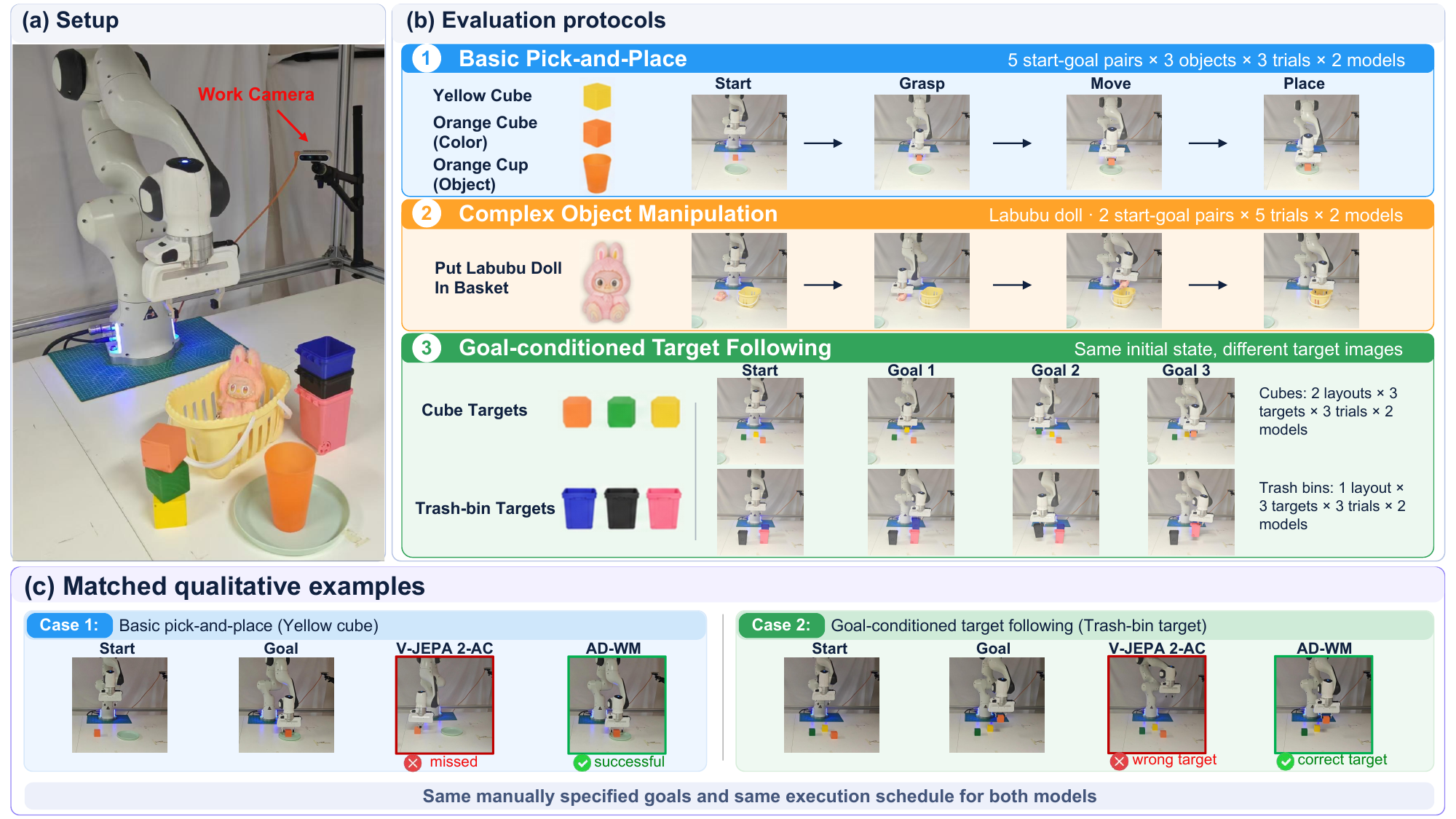}
  \caption{\textbf{Real-robot evaluation.} (a) Franka setup with a single camera. (b) Basic pick-and-place, complex-object, and specified-target protocols use 45, 10, and 27 trials per model. (c) Matched examples illustrate grasping and target selection. Both models share image goals, execution stages, gripper handling, and action clipping.}
  \label{fig:robot}
  \vspace{-5mm}
\end{figure*}

\subsection{Planning Diagnostics}
\label{sec:diagnostics}

\textbf{Evaluation setup.}
To answer Q3, Table~\ref{tab:diagnostic-summary} compares prediction error, candidate selection, and HS. Factual errors use 61,999 held-out clips with three-frame context. Candidate selection evaluates five default-weight variants over three seeds on 64 shared cases with 300 candidates each. Banks contain four reference sequences (expert, zero, negated-expert, and reversed-expert) and 99/99/98 near/medium/far perturbations. Actions and simulator endpoints are shared, while costs use each model's encoder. The default elite size $k=30$ matches CEM, with $k=15,60$ testing robustness.

\textbf{Prediction and selection quality.}
LeWM has the lowest factual MSE but also the lowest HS; \method reaches $52.0\%$ HS despite the highest MSE. These errors depend on representation and scale. Across 15 model--seed observations, HS correlates with CAD at $\rho=-0.399$ and with negative $R_{30}$ and $\bar R_{30}$ at $0.863$ and $0.810$, respectively. The regret associations persist at $k=15,60$. Hierarchical paired bootstrap over seeds and cases (20,000 resamples) gives successive $R_{30}$ reductions along LeWM--Res--Res+Inv--\method of $0.028$ [$0.014,0.043$], $0.002$ [$-0.003,0.009$], and $0.014$ [$0.009,0.023$] (95\% CIs). The Inv-only increment remains unresolved; Res+MI and \method have similar mean regret. From LeWM to \method, realized-best candidate retention falls from $0.198$ to $0.135$; a nearly tied alternative can still yield low regret. These associations concern fixed-bank selection, not adaptive CEM trajectories.

\textbf{Local dynamics and search geometry.}
We examine the first five steps of 25-step rollouts and 20 CEM landscapes on a $31\times31$ grid along the first two principal directions of the final elites. From LeWM to Res, latent-increment MSE decreases from $0.0067$ to $0.0055$, and the fraction of consecutive predicted increments with negative cosine similarity falls from $0.242$ to $0.146$. From LeWM to \method, the distance between the CEM center and the lowest-cost grid point decreases from $0.827$ to $0.130$, while mean elite-to-center distance falls from $0.448$ to $0.082$. These results are consistent with more stable local updates and more concentrated search.

\subsection{Real-Robot Transfer}
\label{sec:robot}

\textbf{Setup and post-training.}
To address Q4, we evaluate transfer on the Franka setup in Fig.~\ref{fig:robot}(a). We compare \method with V-JEPA~2-AC~\cite{vjepa2} using the same frozen V-JEPA~2 ViT-G encoder, filtered \texttt{cadene/droid\_1.0.1} data~\cite{droid}, predictor architecture, training schedule, CEM planner, clipping, and deployment stack. Only transition parameterization and auxiliary objectives differ. No laboratory images or demonstrations are used for adaptation.
The predictor has dimension 1024, depth 24, and 16 heads. Post-training uses four A800 GPUs, bf16, 315 epochs, 300 iterations per epoch, batch size 8 per GPU, and peak learning rate $4.25\times10^{-4}$. Auxiliary weights are $\lambda_{\mathrm{inv}}=0.1$, $\lambda_{\mathrm{MI}}=5\times10^{-5}$, and $\beta=0.1$, without tuning on Franka evaluation results. Actions are DROID-style 7D Cartesian/gripper deltas. Each replan predicts one latent transition using 800 CEM samples, 10 elites, and 10 iterations.

\textbf{Evaluation protocol.}
A single side/rear RealSense camera provides eight frames at 4 fps with a $256\times256$ crop. MPC follows manually supplied grasp, move, and place image goals, with 10/10/4 execution steps. Cartesian deltas are clipped to norm $0.075$ and converted to absolute base-frame targets. Gripper commands are thresholded, with forced closing during grasp and opening during place. This evaluates short-horizon MPC within a structured subgoal pipeline.

Each model receives 45 basic pick-and-place trials (three objects, five start-goal settings, three repeats), 10 complex-object trials, and 27 specified-target trials. The total is 82 trials per model and 164 overall. No trials are excluded, and safety stops count as failures. Evaluation uses separate, non-randomized model blocks.

\textbf{Results.}
Fig.~\ref{fig:robot}(b) illustrates three protocols: pick-and-place across object colors and shapes, placing a Labubu doll in a basket, and selecting objects through different goal images. Table~\ref{tab:robot} shows higher basic pick-and-place success, from $42.2\%$ to $71.1\%$. All three objects improve ($33.3\%\to60.0\%$, $46.7\%\to73.3\%$, and $46.7\%\to80.0\%$). Complex-object and specified-target results favor \method, though smaller samples limit precision. In Fig.~\ref{fig:robot}(c), \method grasps the cube and selects the requested trash bin, while V-JEPA~2-AC misses the grasp or moves the wrong target. Abnormal motion denotes uncontrolled or implausible movement. Specified-target metrics distinguish moving the requested object from completing lift-and-place. These results support transfer within the shared deployment pipeline.

\begin{table}[t]
\centering
\caption{\textbf{Zero-shot Franka results (count/total).} Abnormal motion counts adverse events. Bold marks better results.}
\label{tab:robot}
\footnotesize
\setlength{\tabcolsep}{3pt}
\renewcommand{\arraystretch}{1.02}
\begin{adjustbox}{max width=\columnwidth}
\begin{tabular}{@{}llrr@{}}
\toprule
\rowcolor{ADHeader}
Protocol & Metric & V-JEPA~2-AC & \textbf{\method} \\
\midrule
Basic & Success $\uparrow$ & 19/45 & \textbf{32/45} \\
\midrule
Complex & Success $\uparrow$ & 2/10 & \textbf{5/10} \\
 & Abnormal motion $\downarrow$ & 6/10 & \textbf{2/10} \\
\midrule
Target & Target moved $\uparrow$ & 14/27 & \textbf{21/27} \\
 & Lift-and-place $\uparrow$ & 9/27 & \textbf{17/27} \\
\bottomrule
\end{tabular}
\end{adjustbox}
\vspace{-5mm}
\end{table}

\section{Conclusion and Limitations}

We present \method, an action-discriminative world model for counterfactual MPC. It combines residual prediction with predictor-level action recovery to preserve action information without changing the planner. Experiments show gains over matched LeWM in four of five simulation environments and transfer to Franka without lab-specific adaptation. Cube ablations identify residual prediction and MI as the largest contributors, while Inv has a smaller, weight-dependent effect on mean success. Cube diagnostics show that elite regret tracks success more closely than factual error or whole-bank ranking.

Diagnostics remain limited to Cube, and Scene's gain is unresolved over three seeds. External comparisons differ in inference and checkpoint counts. Simulation and robot post-training use different auxiliary weights. Robot evaluation uses manual subgoals and non-randomized trials with one site, camera, and backbone. Future work examines broader settings and reduces reliance on manual subgoals.

\bibliographystyle{IEEEtran}
\bibliography{references}

\end{document}